%% file: main.tex
\documentclass[runningheads]{llncs}

\usepackage{eccv}

\usepackage{eccvabbrv}

\usepackage{graphicx}
\usepackage{booktabs}
\usepackage{multirow}
\usepackage{bm}
\usepackage{makecell}
\usepackage[accsupp]{axessibility}  
\usepackage[misc]{ifsym}
\usepackage{amssymb}

\newcommand{\method}{MAG-3D\xspace}
\newcommand{\doubao}{Seed-1.6\xspace}
\newcommand{\methodgpt}{MAG-3D$_\text{GPT-4o}$\xspace}
\newcommand{\methoddb}{MAG-3D$_\text{Seed-1.6}$\xspace}
\newcommand{\visualmem}{3D Visual Memory\xspace}

\newif\ifshowtodos
\showtodostrue   

\usepackage[table]{xcolor}
\newcommand{\gain}[1]{\textcolor{green!50!black}{\scriptsize$\uparrow$#1}}

\usepackage{etoc}
\providecommand{\authcount}[1]{} 
\etocsetlevel{title}{6}         
\etocsetlevel{author}{6}

\usepackage{hyperref}

\usepackage{orcidlink}

\begin{document}

\title{MAG-3D: Multi-Agent Grounded Reasoning for 3D Understanding} 

\titlerunning{MAG-3D: Multi-Agent Grounded Reasoning for 3D Understanding}

\makeatletter
\newcommand{\eqmark}{\text{\@fnsymbol{1}}} 
\newcommand{\corrmark}{\text{\Letter}}     
\makeatother

\author{Henry Zheng\inst{1,2,\eqmark} \and
Chenyue Fang\inst{1,\eqmark}\and
Rui Huang\inst{1,2} \and 
Siyuan Wei\inst{2} \and 
Xiao Liu\inst{2} \and 
Gao Huang\inst{1,\corrmark}}
\authorrunning{H.~Zheng et al.}

\institute{Tsinghua University \and Pico, ByteDance \\
\email{
\{jh-zheng22,fcy23,hr20\}@mails.tsinghua.edu.cn,\\
\{weisiyuan.buaa,liuxiao.ai\}@bytedance.com,\\
\corrmark~gaohuang@tsinghua.edu.cn}
}

\maketitle

\begingroup
\renewcommand\thefootnote{}
\footnotetext{\textsuperscript{\eqmark} Equal contributions. ~~~\corrmark\ Corresponding author.}
\endgroup

\label{sec:intro}
\begin{figure}[h]
    \centering 
    \includegraphics[width=1.0\textwidth]{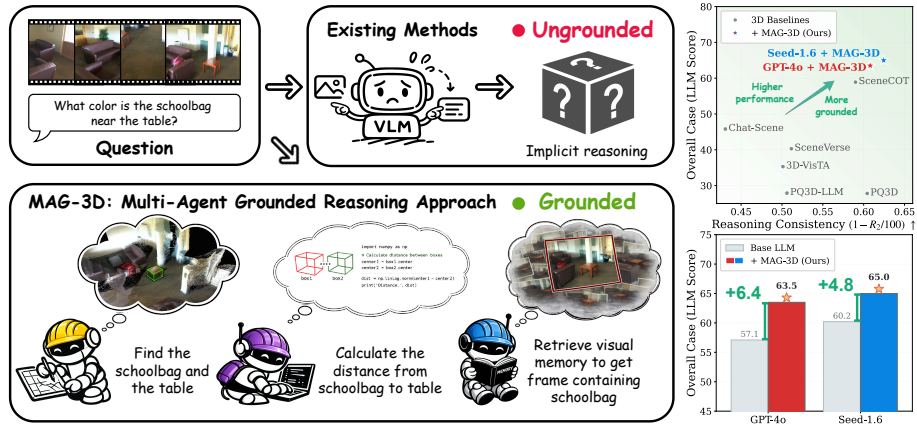}
    \caption{
    \textbf{Comparison between existing methods and MAG-3D.} While existing methods (top) often rely on implicit reasoning, MAG-3D (bottom) introduces a training-free multi-agent framework for grounded 3D reasoning. Crucially, MAG-3D avoids in-domain tuning or hand-crafted pipelines, offering superior flexibility to adapt to diverse queries and environments by dynamically coordinating expert agents. As shown on the right, this approach significantly boosts reasoning consistency and accuracy, achieving state-of-the-art performance with off-the-shelf VLMs. }
    \label{fig:teaser}
    \vspace{-25pt}
\end{figure}
\input{sec/0abstract}
\input{sec/1intro}
\input{sec/2relatedworks}
\input{sec/3methods}
\input{sec/4experiments}
\input{sec/5conclusion}

\clearpage

\bibliographystyle{splncs04}
\bibliography{main}

\input{sec/appendix}

\end{document}

%% file: sec/0abstract.tex
\begin{abstract}
Vision-language models (VLMs) have achieved strong performance in multimodal understanding and reasoning, yet grounded reasoning in 3D scenes remains underexplored. Effective 3D reasoning hinges on accurate grounding: to answer open-ended queries, a model must first identify query-relevant objects and regions in a complex scene, and then reason about their spatial and geometric relationships. Recent approaches have demonstrated strong potential for grounded 3D reasoning. However, they often rely on in-domain tuning or hand-crafted reasoning pipelines, which limit their flexibility and zero-shot generalization to novel environments. In this work, we present MAG-3D, a training-free multi-agent framework for grounded 3D reasoning with off-the-shelf VLMs. Instead of relying on task-specific training or fixed reasoning procedures, MAG-3D dynamically coordinates expert agents to address the key challenges of 3D reasoning. Specifically, we propose a planning agent that decomposes the task and orchestrates the overall reasoning process, a grounding agent that performs free-form 3D grounding and relevant frame retrieval from extensive 3D scene observations, and a coding agent that conducts flexible geometric reasoning and explicit verification through executable programs. This multi-agent collaborative design enables flexible training-free 3D grounded reasoning across diverse scenes and achieves state-of-the-art performance on challenging benchmarks.

\keywords{3D Vision-Language Understanding \and Multi-Agent Framework \and Grounded 3D Reasoning}
\end{abstract}

%% file: sec/1intro.tex
\section{Introduction}

Vision-language models (VLMs) have rapidly evolved into powerful multimodal systems with strong capabilities in visual understanding and reasoning, achieving impressive performance on diverse tasks such as open-ended image and video question answering, GUI grounding, and visual code generation~\cite{liu2023visual,openai2024gpt4technicalreport,hong2024cogagent,si2025design2code}. However, extending these capabilities to embodied spatial reasoning within 3D scenes remains challenging. Success on conventional 2D vision-language tasks does not directly imply reliable reasoning over 3D geometry, spatial relations, and scene structure from visual observations~\cite{chen2024spatialvlm,cheng2024spatialrgpt,zhu25llava3d,chen2024ll3da}.

Effective 3D reasoning inextricably depends on accurate grounding.
Answering open-ended queries in a 3D scene requires
identifying task-relevant entities, aggregating evidence from fragmented observations, and performing consistent spatial deduction.
This is fundamentally different from 2D settings, where language priors or surface-level semantic correlations often suffice. 
Without explicit grounding in the 3D scene structure, VLMs are prone to hallucinations, generating plausible-sounding responses that are factually disconnected from the physical environment.
Thus, bridging the gap between high-level semantic reasoning and low-level geometric grounding is essential.

Recent efforts have made strides toward grounded 3D reasoning but face distinct limitations. Reasoning-oriented methods~\cite{ouyang2025spacer, sensenova-si, yang2025visual, liu2025spatial, chen2025think, linghu2026scenecot} typically enhance performance through instruction tuning on 3D data. However, they rely heavily on specialized supervision and in-domain adaptation, limiting their generalization to unseen scenes.
Alternatively, tool-augmented methods~\cite{han2025tiger,luo2026pySpatial} leverage external perception modules for evidence collection. While promising, they often depend on rigid, predefined reasoning pipelines or static tool configurations.
Such designs lack the flexibility required for open-world settings, where diverse user queries necessitate dynamic and adaptive reasoning paths.

To address these challenges, we present \method, a training-free, multi-agent framework that empowers off-the-shelf VLMs to perform robust \emph{grounded 3D reasoning}.
Departing from static pipelines, \method employs a central \emph{Planning Agent} to dynamically decompose complex queries and coordinate the reasoning process. 
Specifically, we introduce two agents for robust, open-world-compatible 3D grounded reasoning: (1) a \emph{Grounding Agent} that resolves free-form language references to localize query-relevant objects and retrieve supporting views from extensive scene observations, forming a shared visual memory; and (2) a \emph{Coding Agent} that performs explicit geometric computation and verification by generating task-specific executable programs conditioned on the task query and the accumulated scene state. This design yields a reasoning process that is flexible, interpretable, and explicitly grounded in the 3D scene.

Our experiments on the Beacon3D and the MSQA (Multi-modal Situated Question Answering) tasks show that \method consistently achieves state-of-the-art performance on open-ended 3D question answering without additional training. On Beacon3D, \method improves its corresponding base planners substantially (\eg, +6.4 and +3.2 over pure GPT-4o on case-level and object-level QA Scores, respectively), and \methoddb achieves the best overall performance among prior methods, outperforming the previous strongest approach SceneCOT~\cite{linghu2026scenecot} by +6.1 points in case-level QA Score and +4.3 in the stricter object-level score. Without training on the MSQA domain, \method still achieves the best performance among zero-shot methods under both the official and vision-only settings. Finally, on Beacon3D coherence evaluation, \method achieves the highest \emph{Good Coherence} metric, further supporting that the accuracy gains stem from more faithful grounded reasoning.

In summary, our contributions are as follows:
\begin{itemize}
    \item We propose \method, a training-free multi-agent framework that enables grounded 3D reasoning with off-the-shelf VLMs, avoiding task-specific training or in-domain tuning.
    \item We develop an agentic design that explicitly tackles 3D reasoning via coordinated planning, open-vocabulary localization, and geometric verification.
    \item We validate \method on Beacon3D and MSQA, achieving state-of-the-art performance among training-free methods on both benchmarks; on Beacon3D, our grounding--QA analysis also shows the strongest grounding--QA coherence, indicating improved grounding--QA alignment.
\end{itemize}

%% file: sec/2relatedworks.tex
\section{Related Work}
\subsection{VLMs for 3D Spatial Understanding}
The 3D spatial understanding task is a non-trivial capability fundamental to embodied intelligence, necessitating long-horizon multi-view integration, explicit modeling of spatial relations, and globally consistent reasoning. Recent evidence suggests that Large Vision-Language Models (LVLMs) remain brittle when spatial demands extend beyond single-view 2D recognition \cite{yang2024think}. To address such 3D-centric challenges, recent 3D-VLMs explore diverse scene representation strategies. One prevalent direction leverages multi-view 2D images as a proxy, either by augmenting the input representations with 3D information \cite{zhu25llava3d, zheng24video3d, qi25gpt4scene} or by introducing novel 3D-aware training objectives \cite{wang25ross3d}. Alternatively, instance-aware strategies parse the scene into discrete objects via 3D semantic instance segmentation \cite{Schult23mask3d}, bridging these object-level representations to LLMs through advanced multimodal fusion \cite{deng253dllava, yu25inst3d}. For specialized tasks like object localization, some approaches forgo general-purpose LLMs entirely in favor of self-supervised learning on point clouds \cite{arnaudmcvay25locate3d}. Beyond architectural variations, another line of research focuses on enhancing internal spatial representations through specialized training on reasoning-intensive data \cite{linghu2026scenecot, ouyang2025spacer} or the construction of large-scale 3D Question-Answering (QA) datasets \cite{linghu2024multi, zhang2025from, fan2025vlm3rvisionlanguagemodelsaugmented}. However, both architectural and data-driven methods remain constrained by the scarcity and high cost of 3D annotations, often struggling to generalize to the unstructured complexities of open-world environments. Consequently, achieving robust, multi-step, and globally consistent reasoning remains a primary research focus in the field.

\subsection{Tool-Grounded Reasoning}
Building on the limitations discussed earlier, a promising direction for improving spatial reasoning is to integrate external expertise via specialized tools. DeepSeek introduces a thinking retention mechanism for tool calling, retaining tool invocations and their returned observations to support multi-step tool-augmented reasoning \cite{deepseekai2025deepseekv32pushingfrontieropen}. More broadly, such designs follow the general line of tool-using and memory-aware agent frameworks for interleaving reasoning with actions and managing intermediate context \cite{yao2023react, schick2023toolformerlanguagemodelsteach}.
In 2D-VLMs, tool usage is often framed as a cycle of active perception, where models iteratively “observe–think–act” \cite{wang25simpleo3, wu23vguidedvisualsearch, hu24visualsketchpad, zheng25deepeyes, Gupta2022VisProg, su2025openthinkimg, vstar, li2025dyfotrainingfreedynamicfocus}. Extending this framework to video, tool-augmented LVLMs have been developed to externalize spatiotemporal evidence through modules such as tracking, temporal grounding, and video segmentation \cite{zhang2025thinking, yuan2025thinkvideosagenticlongvideo, fan2025toolaugmentedspatiotemporalreasoningstreamlining, zhang2025deep, fan2025videoagent}. These modules help capture visual scenes, enabling models to reason across spatial dimensions in 2D and both spatial and temporal dimensions in video. Towards 3D understanding, recent efforts also employ tool-augmented reasoning that maps semantic queries and 2D observations into an explicit geometric state amenable to computation and exploration \cite{han2025tiger, chen2025geometrically, luo2026pySpatial}. However, these approaches remain limited by either reliance on additional tuning on in-domain data or relatively hand-crafted reasoning schemes that can reduce expressiveness and robustness in open-world settings. Motivated by these limitations, we propose \method, a flexible training-free multi-agent framework for 3D reasoning with a grounded verifiable multi-modal reasoning chain.

%% file: sec/3methods.tex
\section{Methods}
This section presents our training-free multi-agent framework for 3D spatial understanding from multi-view RGB observations in diverse settings. We first formalize the task and summarize the overall system design in Sec.~\ref{sec:overview}, then detail the \emph{Planning} (Sec.~\ref{sec:planning}), \emph{Grounding} (Sec.~\ref{sec:grounding}), and \emph{Coding} (Sec.~\ref{sec:coding}) agents, and explain how they interact in an end-to-end reasoning loop.

\subsection{Overview}
\label{sec:overview}
\begin{figure}[t]
    \centering 
    \includegraphics[width=1.0\textwidth]{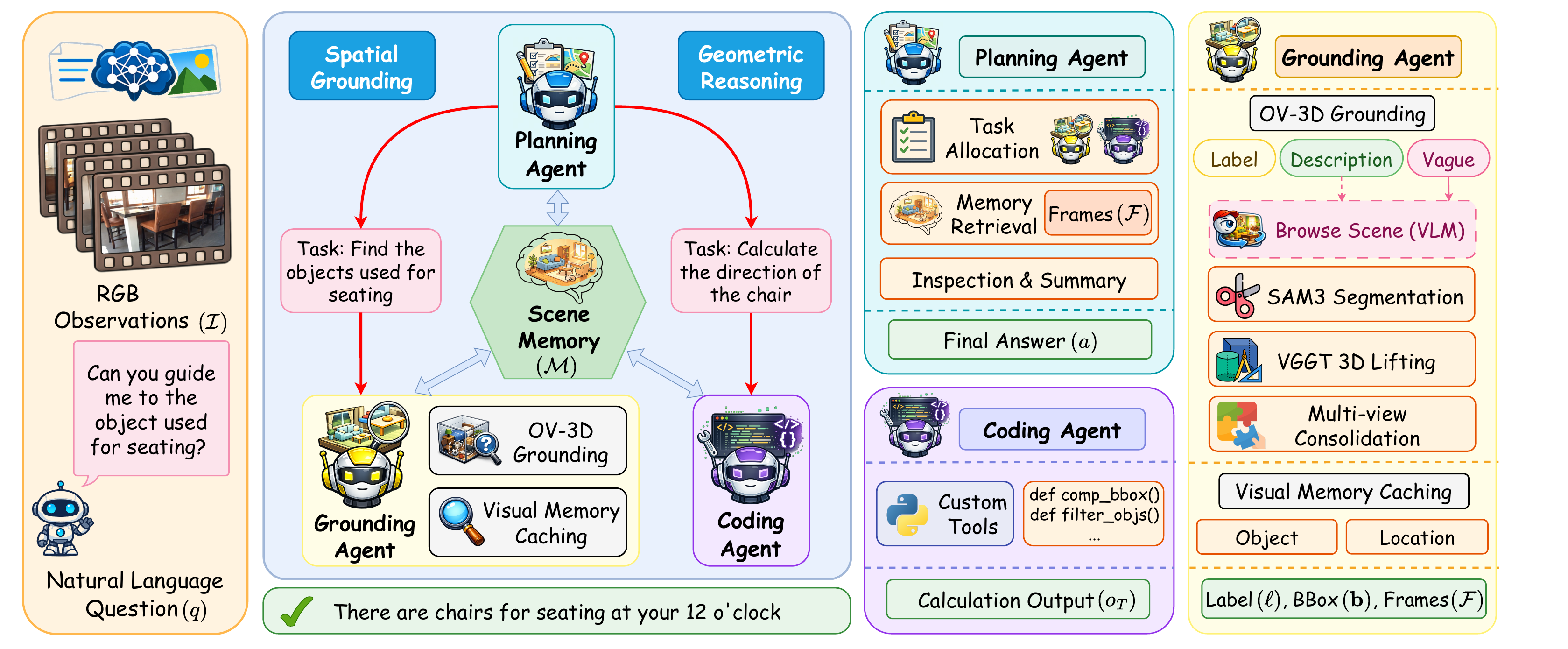}
    \caption{\textbf{The overall framework of \method.} Given a question \(q\) and RGB observations $\mathcal{I}$, the planning agent dynamically orchestrates expert agents for spatial grounding, geometric reasoning, and retrieves information from scene memory $\mathcal{M}$ as needed. The framework maintains explicit, inspectable intermediate results (\eg, grounded instances and geometric results), which are then aggregated and summarized into the final answer \(a\). The right panel summarizes the functional components of each agent.}
    \label{fig:framework}
\end{figure}

We study open-world 3D spatial understanding from video or multi-view RGB observations, optionally with point clouds \(\mathcal{P}\). Let \(\mathcal{I} = (I_i)_{i=1}^{N}\) denote the RGB observations, where \(I_i\) is the \(i\)-th frame or view. Given \(\mathcal{I}\), optional \(\mathcal{P}\), and a natural-language query \(q\), we predict an answer \(a\) grounded in the underlying 3D scene. In contrast to prior 3D-VLM systems~\cite{qi25gpt4scene, wang25simpleo3, zheng24video3d} that depend on explicit 3D priors (closed-vocabulary 3D detectors) or task-specific fine-tuning, our approach is training-free and operates with off-the-shelf image-level foundation models. We decompose grounded 3D reasoning into three collaborating agents: a \emph{Planning Agent} that decomposes \(q\) and orchestrates the process, a \emph{Grounding Agent} that performs free-form 3D grounding by discovering query-relevant objects or regions and associating them with supporting observations as visual memory, and a \emph{Coding Agent} that translates spatial-language intents into executable geometric programs for explicit computation and verification. The agents communicate via a shared scene memory \(\mathcal{M}\), which stores intermediate results (\eg, object candidates, reconstructed geometry, visual memory, and measurements), enabling iterative grounding-verification for a grounded answer (Fig.~\ref{fig:framework}).

\subsection{Planning Agent for Multi-agent Orchestration}
\label{sec:planning}
The Planning Agent serves as the central coordinator of the framework. Given the query \(q\) and the current state of the shared scene memory \(\mathcal{M}\), it decomposes the task into sub-goals, allocates each sub-goal to the appropriate agent, and decides whether additional grounding or geometric reasoning is required. Rather than relying on fixed human-crafted rules or predefined reasoning pipelines, the Planning Agent enables adaptive coordination, allowing the framework to handle diverse 3D question types in a flexible yet generalizable manner. After each step, intermediate outputs are written back to \(\mathcal{M}\), enabling it to update the reasoning state and refine subsequent actions. If intermediate results are ambiguous, incomplete, or inconsistent, it can re-plan, re-invoke agents to retry failed steps with updated evidence. Once sufficient verified evidence has been collected, it summarizes the intermediate results and produces the final grounded answer.
\subsection{Grounding Agent for Adaptive 3D Grounding}
\label{sec:grounding}
\subsubsection{Open-Vocabulary 3D Grounding.}
Given a natural-language referential query, the Grounding Agent localizes the referenced object as a set of grounded instances
$\{(\ell,\mathbf{b})\}$, where $\ell$ is an open-vocabulary label and $\mathbf{b}$ a 3D bounding box. In this work, we treat grounding as
\emph{open-vocabulary 3D instance identification} and decouple it from downstream spatial reasoning: the agent only localizes query-relevant object instances, while geometric or spatial relations are handled by downstream explicit geometric reasoning modules.

We support both \emph{clear} references (explicit categories or functional/attribute descriptions (\eg, \emph{``chair''}, \emph{``objects for storing food''})) and \emph{vague} references (\eg, \emph{``what is to the left of the chair?''}) specified implicitly by context or relations. For clear prompts, we directly use the input text as the grounding cue. For vague prompts (or when initial grounding fails), we provide \emph{Browse Scene}, whereas the VLM could utilize to propose a small set of candidate labels from RGB observations, which are then used as grounding prompts.

Our grounding is training-free, combining 2D open-vocabulary instance segmentation with multi-view geometric lifting and consolidation. We obtain 2D instance masks with SAM3~\cite{carion2025sam3segmentconcepts} and lift them to 3D using VGGT~\cite{wang2025vggt} using predicted depth and camera poses. Multi-view lifting introduces cross-view inconsistency under occlusion, partial visibility, and noisy pseudo-geometry; we address this with lightweight post-processing: (1) \textbf{visual re-prompting} resolves mask conflicts arising from overlap by re-querying the disputed region together with both competing labels and keeping the most consistent hypothesis~\cite{huang2026openvoxeltrainingfreegroupingcaptioning}; (2) \textbf{label-gated geometric merging} removes cross-view duplicates by \emph{overlap-based fusion}: for instances with the same predicted label, we compute a containment-based overlap score on voxelized points,
$s_\text{o}(A,B)=\max\!\left(\frac{|V_A\cap V_B|}{|V_A|},\frac{|V_A\cap V_B|}{|V_B|}\right)$, where $V_A$ and $V_B$ denote voxelized 3D inputs;
we merge instances when $s_\text{o}(A,B)\ge\tau$ (union of their 3D supports), where $\tau$ is a set threshold; and (3) \textbf{yaw-oriented box fitting} fits a ground-aligned 3D box by projecting the merged points to bird's-eye view and computing the minimum-area enclosing rotated rectangle (yaw only). More implementation details are provided in the appendix.

\subsubsection{\visualmem.}
To reduce irrelevant visual context while preserving occlusion-aware evidence, we maintain a \emph{3D Visual Memory} built from VGGT reconstructions~\cite{wang2025vggt}. For each frame $i$, we store its RGB image and point map $\mathbf{P}^{(i)}\in\mathbb{R}^{H_p\times W_p\times 3}$ in the scene coordinate system, providing persistent 2D--3D correspondences for later inspection.

\paragraph{Visual Memory Caching.} When the Grounding Agent produces a 3D instance box $\mathbf{b}$, we cache a compact visual memory entry by ranking frames according to how much \emph{3D volume} of the queried instance region is actually observed (\ie, overlapped) by the frame’s pixel-aligned point map. Concretely, given a 3D query region $\mathcal{R}\subset\mathbb{R}^3$, we define the per-frame 3D coverage score $f_i(\mathcal{R})$ as the \emph{occupied volume} induced by points from frame $i$ that fall inside $\mathcal{R}$, and cache the top-$K$ frames with the highest scores:
\[
\mathcal{F}(\mathcal{R})=\operatorname{TopK}_{i\in\{1,\dots,N\}}\, f_i(\mathcal{R}),\qquad
f_i(\mathcal{R})=\Delta^3\left|\left\{\mathcal{V}_\Delta(\mathbf{p})~\middle|~
\mathbf{p}\in \mathbf{P}^{(i)}\cap \mathcal{R}\right\}\right|.
\]
Here $\mathcal{V}_\Delta(\cdot)$ maps a 3D point to its voxel index under grid resolution $\Delta$, so $f_i(\mathcal{R})$ measures the 3D \emph{volume} of $\mathcal{R}$ covered by image-overlapping points rather than raw point counts. We cache the indices in $\mathcal{F}(\mathcal{R})$ together with the corresponding point maps, yielding an instance-centric cache that can be reused across multiple downstream queries without repeated global scanning.

\paragraph{Visual Memory Retrieval.}
Downstream modules retrieve visual evidence by instantiating a 3D region $\mathcal{R}$ and returning the top-$K$ frames ranked by the volumetric coverage score $f_i(\mathcal{R})$, along with their 2D--3D mappings. The $\mathcal{R}$ is instantiated either as the axis-aligned region of a grounded box $\mathbf{b}$ (instance retrieval) or as a small query-aligned local volume around a target location and heading (location retrieval; see appendix for the region definition and parameters). Compared to 2D-only caching that ranks frames by instance mask area, volumetric 3D coverage prefers viewpoints that observe a larger portion of the underlying 3D region, making retrieval more robust to partial observations across views.

\subsection{Coding Agent for Precise Calculations}
\label{sec:coding}
The Coding Agent executes grounded geometric reasoning for tasks that are difficult to solve reliably by direct language generation. Given the planning agent's coding instruction and scene memory $\mathcal{M}$, containing information such as 3D bounding boxes, images, and question specific ego-poses, it generates executable Python code to perform task-specific computation and return the final result.

The agent interacts with a Python interpreter in a multi-round manner. At each step, it proposes python codes, executes it, and uses the result as feedback for possible revision. Formally, given a coding query \(q_\text{c}\) and scene memory $\mathcal{M}$, the agent iteratively produces an action and receives execution feedback,
\[
c_t = \pi^t_{\mathrm{code}}(q_\text{c}, \mathcal{M}), \quad o_t = \mathcal{E}(c_t),\quad t=1,\dots,T,
\]
where $\mathcal{E}(\cdot)$ denotes the Python interpreter, $T$ is the step when the agent terminates, and the final result is $o_T$. This process allows the agent to verify intermediate computations through actual execution, which improves robustness for multi-step spatial reasoning. A few built-in function libraries are available for common calculations, and other cases will be handled with custom Python code. Additional details are given in the appendix.

%% file: sec/4experiments.tex
\section{Experiments}
\label{sec:experiments}
In this section, we outline the experimental setup (Sec.~\ref{sec:experimental_setup}), benchmarks, and metrics, including QA Score and Beacon3D GQA-Chain coherence (Sec.~\ref{sec:bench_eval}), then present Beacon3D and MSQA results (Sec.~\ref{sec:quantitative_results}). Finally, we provide ablations (Sec.~\ref{sec:ablation}) and qualitative analyses with intermediate evidence (Sec.~\ref{sec:qualitative_results}).

\subsection{Experimental Setup}
\label{sec:experimental_setup}
For the main experiments, we instantiate the planning agent with either \doubao (\texttt{doubao-seed-1-6-250615}) or GPT-4o (\texttt{gpt-4o-0806}). Unless otherwise specified, all other agents are instantiated with \doubao throughout the experiments. We adopt \doubao as the default backbone for the sub-agents because it demonstrates strong and stable performance on agentic tasks within our framework, enabling a controlled study of the effect of different planner models.

Under this configuration, we evaluate our method on Beacon3D~\cite{huang25unveiling} and MSQA~\cite{linghu2024multi}. We primarily follow the official MSQA setting to ensure fair comparison with prior methods. In addition, we include a vision-only MSQA setting as a supplementary study to examine robustness when only RGB videos are available, a setting that is also relevant for real-world implementations. 

For the ablation studies, unless otherwise specified, we use \doubao as the backbone for all agents and conduct experiments on the vision-only setting of the MSQA ScanNet benchmark. In the ablation, we also report a single-agent tool-use baseline, where a single VLM, prompted with a fixed tool-calling template, invokes the same grounding module and a predefined task-specific tool set. 

\subsection{Benchmarks and Evaluation Metrics}
\label{sec:bench_eval}

We evaluate on challenging benchmarks that require models to understand 3D inputs and generate open-ended answers, thereby reducing shortcuts commonly introduced by multiple-choice evaluation. This setting is particularly important for trustworthy 3D scene understanding, where answers should be grounded in the corresponding relevant objects and scene context rather than driven by dataset bias or spurious correlations. To this end, we adopt Beacon3D~\cite{huang25unveiling} as our primary benchmark, since it evaluates not only QA performance but also grounding-QA coherence through Grounding-QA Chains (GQA-Chains). As a supplementary evaluation, we additionally report results on MSQA, which measures multimodal situated 3DQA performance.

\subsubsection{Question Answering Evaluation (QA Score).}
Following Beacon3D~\cite{huang25unveiling} and MSQA~\cite{linghu2024multi}, we evaluate answer quality using an LLM-based judge. For each QA pair, the judge assigns a score $M_i \in \{1,2,3,4,5\}$. Then, the case-level QA Score, measuring the average quality of individual QA pairs, is computed as
\begin{equation}
\mathrm{QA\mbox{-}Score}_{\mathrm{Case}}
=
\frac{1}{|\mathcal{Q}|}
\sum_{i \in \mathcal{Q}}
\frac{M_i - 1}{4}
\times 100,
\end{equation}
where $\mathcal{Q}$ is the set of QA pairs. For Beacon3D, we provide additional object-level QA Scores,
\begin{equation}
\mathrm{QA\mbox{-}Score}_{\mathrm{Obj}}
=
\frac{1}{|\mathcal{O}|}
\sum_{o \in \mathcal{O}}
\mathbf{1}\!\left[\min_{k=1,2,3} M_{o,k} \ge 4\right]
\times 100,
\end{equation}
where $\mathcal{O}$ denotes the set of objects and $M_{o,k}$ is the judge score for the $k$-th QA pair associated with object $o$. The object-level metric requires all three QA pairs for an object to receive sufficiently high scores. 

\subsubsection{Grounding--QA Coherence (GQA-Chains).}
Beacon3D evaluates grounding--QA coherence by categorizing each example into four GQA-Chain outcomes:
Good coherence $G$ $(\checkmark,\checkmark)$,
Type-1 broken coherence $T_1$ $(\checkmark,\times)$,
Type-2 broken coherence $T_2$ $(\times,\checkmark)$,
and Double failure $D$ $(\times,\times)$, where the tuple denotes (grounding, QA).
In addition to reporting the percentage of examples in each outcome, Beacon3D defines two conditional ratios:
\begin{equation}
R_1 = \frac{T_1}{T_1 + D}\times 100,
\qquad
R_2 = \frac{T_2}{T_2 + G}\times 100.
\end{equation}
$R_1$ measures the fraction of QA failures that occur despite correct grounding, and $R_2$ measures the fraction of QA successes achieved without correct grounding (shortcut ratio). Lower values indicate better grounding--QA alignment.

\subsection{Quantitative Results}
\label{sec:quantitative_results}

\subsubsection{Results on Beacon3D.}
\input{tables/beacon3d}

Table~\ref{tab:beacon3d_eval} shows that \method consistently improves over its corresponding baseline under both backbones, confirming the effectiveness of our training-free multi-agent design for grounded 3D reasoning. Relative to pure GPT-4o, equipping with \method, \methodgpt improves the overall case-level QA score by +6.4 and the strict object-level QA score by +3.2. Relative to \doubao, \methoddb further improves the case-level metric by 4.8 points and the object-level score by 4.3 points, yielding the best performance. 

We also compare against SceneCOT~\cite{linghu2026scenecot}, which is trained on specially crafted data for 3D grounded reasoning, and observe that \method still achieves higher overall performance without in-domain training. In particular, relative to the previous best method, SceneCOT, in Table~\ref{tab:beacon3d_eval}, our \methoddb improves \emph{Case} by +6.1 and \emph{Obj.} by +4.3 (from 23.2 to 27.5). The improvement on the stricter object-level metric is especially meaningful: since \emph{Obj.} requires all QA pairs associated with an object to be answered reliably, the consistent gains indicate more stable and coherent object-centric reasoning rather than isolated correct answers. Overall, these results establish \method as a strong recipe for zero-shot grounded 3D reasoning with off-the-shelf VLMs. We next analyze whether these gains also translate into better grounding-QA coherence and consistency.

\subsubsection{Grounding-QA Chain Analysis.}
\input{tables/coherence}

Beyond answer accuracy, we further ask whether the model arrives at the correct answer for the \emph{right reason}. To this end, Table~\ref{tab:coherence} and Figure~\ref{fig:grounding_qa_coherence} analyze Grounding-QA Chains on Beacon3D, measuring whether correct answers are supported by correct grounding. The detailed breakdown in Table~\ref{tab:coherence} further supports this conclusion. \methodgpt achieves the lowest Type-1 broken coherence, reducing it by 5.8 relative to the best prior baseline, while \methoddb achieves the lowest double-failure rate at 17.2. 

The clearest trend in Figure~\ref{fig:grounding_qa_coherence} is that \method-based methods substantially increase \emph{Good Coherence}—the proportion of examples in which both grounding and QA are correct—while also reducing incoherent prediction patterns. Compared with the best prior baseline, SceneCOT\cite{linghu2026scenecot} that is trained on in-domain grounded-QA chain-of-thought data, in Figure~\ref{fig:grounding_qa_coherence}, \methodgpt improves \emph{Good Coherence} by 3.2 (from 34.7 to 37.9) and \methoddb improves it by 5.0 (from 34.7 to 39.7). Remarkably, as compared to the best prior non-reasoning baseline, SceneVerse\cite{jia2024sceneverse}, \methoddb improves 19.2 (from 20.5 to 39.7). Moreover, \methoddb further reduces $R_2$ by 3.5 relative to the previous method, indicating fewer cases in which the model produces seemingly correct answers without reliable grounding. Although certain coherence sub-metrics are influenced by the overall QA success rate and are therefore not directly comparable in isolation, the overall trend is clear: explicit multi-agent coordination yields substantially better alignment between grounding and reasoning, rather than merely stronger language-only answer generation.
\subsubsection{Results on MSQA.}

\input{tables/msqa}

Table~\ref{table:msqa_eval} shows that \method also generalizes well to MSQA and achieves state-of-the-art performance under both settings among the zero-shot methods. Under the official setting, \method consistently improves over the corresponding base models, raising the overall score from 3.3 with GPT-4o and 6.4 with \doubao. These results indicate that the advantages of \method are not specific to Beacon3D, but generalize to a different 3DQA benchmark.

As a supplementary robustness study, we additionally evaluate a vision-only setting in which stronger structured cues are removed. Under this setting, both methods improve compared to base models, with \methoddb improving from 29.6 to 42.4, a remarkable gain of 12.8. The gain for the \doubao-based variant suggests that the proposed open-world-targeted grounding agent powered by VGGT\cite{wang2025vggt} is especially helpful when explicit 3D inputs are unavailable. Overall, these results further demonstrate that \method provides a robust solution for grounded 3D reasoning across different 3DQA evaluation settings.

\subsection{Ablation Studies}
\label{sec:ablation}
\input{tables/ablation_framework}

\subsubsection{Component Ablation.} 
Table~\ref{tab:ablations_module_core} reports ablations on the MSQA ScanNet subset. The first row is a single model baseline that achieves a QA Score of 37.5. The second row is a \emph{single-agent tool-use} (T.U.) baseline in Sec~\ref{sec:experimental_setup}, which improves performance to 44.6 and highlights the benefit of explicit grounding and geometric computations provided by tool interaction. However, this design still falls short of our multi-agent system, where a Planning Agent delegates instance localization to the Grounding Agent and geometric computation to the Coding Agent, further increasing QA Score from 44.6 in tool-use baselines to 47.6.

\subsubsection{Agent Backbone Ablation.} We further study how the choice of backbone models affects the grounding and coding agents. Table~\ref{tab:ablations_module_llm} compares various backbone combinations for these two agents. Swapping the coding-agent backbone while keeping GPT-4o-mini~\cite{hurst2024gpt} for grounding leads to only a modest change. In contrast, switching the grounding-agent backbone from GPT-4o-mini to \doubao yields a larger improvement under both coding backbones, reaching 47.0 with Qwen3-Coder~\cite{qwenteam2025qwen3coder} and 47.6 with \doubao. Overall, these results suggest that, within the tested configurations, the grounding agent backbone has a more pronounced impact on the QA Score, while the coding-agent backbone provides smaller gains. A possible explanation is that the code-based verification in 3DQA is relatively lightweight, and thus can already be handled well by modern LLMs that are trained on substantially more complex programming tasks.

\input{tables/ablation_methods}

\subsubsection{Grounding Method Ablation.} 
Table~\ref{tab:grounding_comparison} compares different grounding methods for downstream QA. Without explicit grounding, the model directly answers the question and achieves a QA Score of 37.5. Using instance proposals from the closed-vocabulary 3D instance segmentation model Mask3D~\cite{Schult23mask3d} yields only a small improvement to 38.5. This limited gain suggests that generic closed-vocabulary segmentation is a weak substitute for question-conditioned grounding, as its fixed label space limits its ability to localize the query-relevant objects needed for downstream reasoning. In contrast, our Grounding Agent substantially improves performance to 47.6. This effect is most evident on \emph{Count.} and \emph{Exist.}, which are highly grounding-sensitive since they depend on identifying query-relevant objects. Mask3D improves \emph{Count.} and \emph{Exist.} to 20.9 and 71.9, respectively, whereas our Grounding Agent achieves 27.8 and 78.1, consistent with the corresponding gains in overall QA Score.

\subsubsection{Visual Memory Ablation.} Table~\ref{tab:zoomer_comparison} compares strategies for providing visual context to the grounding agent. \emph{Uniform sampling} is a no-memory baseline where the planner observes uniformly sampled frames, achieving 44.3. A \emph{2D-based} strategy with SAM3\cite{carion2025sam3segmentconcepts} improves to 46.0 by selecting frames that contain query-relevant objects, but it mainly reflects image-space coverage and cannot assess whether the object is fully observed across selected viewpoints. In contrast, our \emph{3D-based visual memory} leverages 3D correspondences to retrieve views that better cover the queried 3D region, obtaining the best QA Score of 47.6, yielding a 3.3 points improvement from the no visual memory baseline.

\subsection{Qualitative Results}
\label{sec:qualitative_results}
\begin{figure}[t]
    \centering 
    \includegraphics[width=1.0\textwidth]{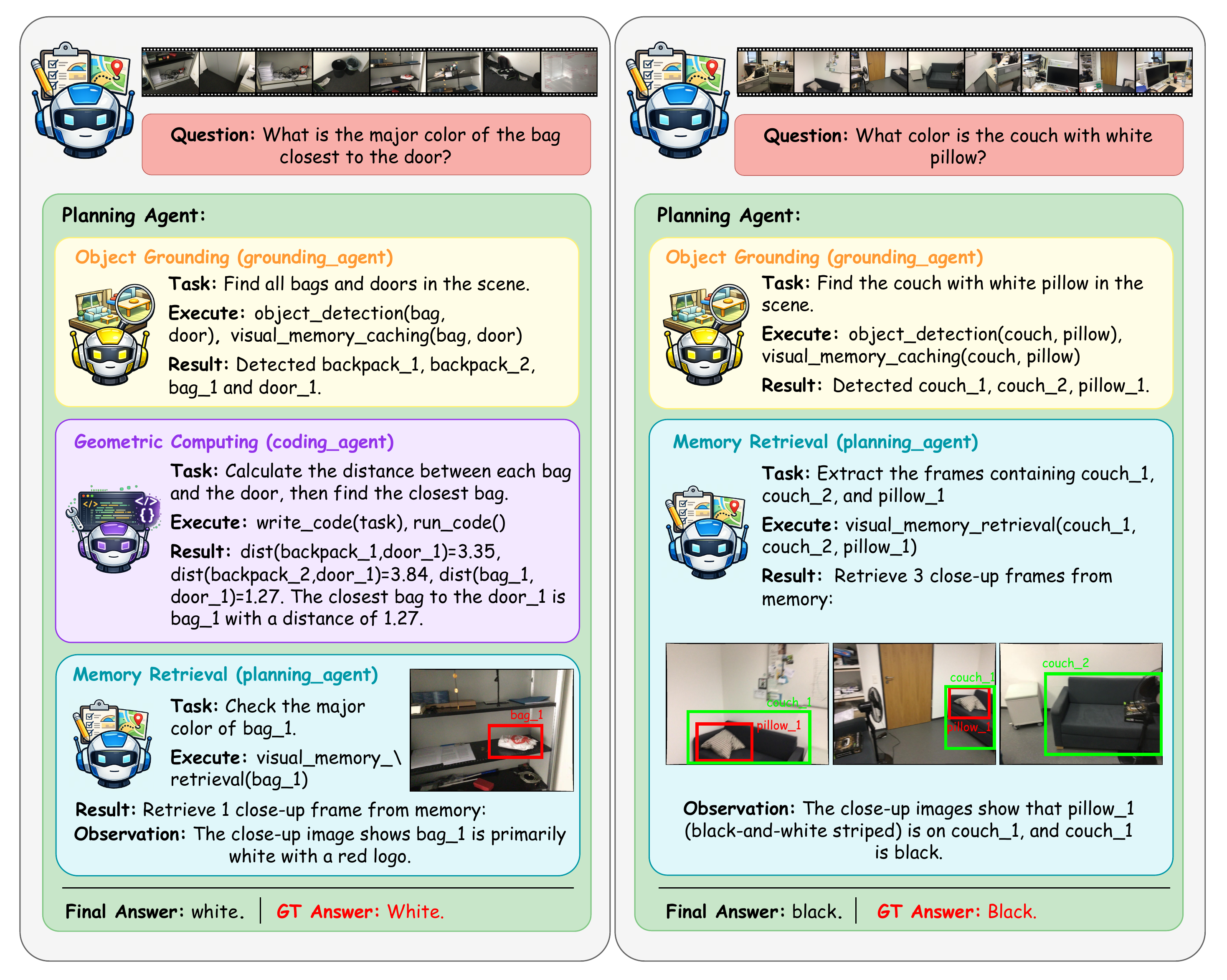}
    \caption{\textbf{Qualitative results on Beacon3D.} For each query, we show (top) the input question and RGB sequence, (middle) intermediate visual and geometric outputs, and (bottom) the predicted and ground-truth answers.}
    \label{fig:examples}
\end{figure}

Fig.~\ref{fig:examples} illustrates how \method answers diverse questions with explicit intermediate outputs. Given a query, \method grounds the relevant objects, performs geometric computation, and retrieves close-up evidence frames from memory to inform the final prediction. Together, these intermediate detections, geometric estimates, and retrieved frames provide structured cues that enable \method to carry out explicit 3D reasoning over objects and their spatial relations.

%% file: tables/beacon3d.tex
\begin{table}[t]
\centering
\setlength{\tabcolsep}{3pt}
\caption{\textbf{Results on Beacon3D}. Results are reported for five question types and two overall QA metrics. ``Case'' and ``Obj''. denote the case-level and object-level QA Scores, respectively. The \textcolor{green!60!black}{green} highlighted rows denote models implemented with our multi-agent method, which consistently improve over the corresponding base models under both planner backbones.}
\begin{tabular}{l|ccccc|cc}
    \toprule
    \multirow{2}{*}{\textbf{Method}} & \multicolumn{5}{c|}{\textbf{Question Type}} & \multicolumn{2}{c}{\textbf{QA Score}} \\
     & \emph{Class} & \emph{App.} & \emph{Geo.} & \emph{Spa.} & \emph{Exi.} & Case & Obj.  \\
    \midrule
    PQ3D-LLM~\cite{huang25unveiling}     & 28.0  & 30.8  & 35.2  & 25.2  & 26.2  & 27.9  & 2.3  \\
    PQ3D~\cite{zhu2024unifying}         & 36.4  & 28.0  & 27.8  & 11.9  & 45.5  & 27.8  & 3.5  \\
    3D-VisTA~\cite{zhu20233d}     & 20.5  & 33.5  & 52.1  & 33.8  & 36.5  & 35.3  & 8.1  \\
    Chat-Scene~\cite{huang2024chat}   & 36.4  & 39.8  & 56.7  & 47.6  & 48.8  & 45.8  & 7.8  \\
    SceneVerse~\cite{jia2024sceneverse}   & 35.6  & 41.7  & 48.9  & 41.9  & 35.7  & 40.3  & 6.6  \\
    LEO~\cite{huang2024embodied}          & 17.4  & 41.0  & 53.2  & 48.7  & 39.7  & 43.2  & 7.8  \\
    SceneCOT~\cite{linghu2026scenecot}          & --  & --  & --  & --  & --  & 58.9  & 23.2  \\
    \midrule
    GPT-4o~\cite{hurst2024gpt}       & 33.3  & 49.9  & 54.9  & 52.1  & 73.8  & 57.1  & 20.2  \\
    \rowcolor{green!4}
    \methodgpt & 29.7  & \textbf{61.2} & \textbf{64.0} & \textbf{65.0} & \textbf{77.1} & \textbf{63.5}\;\gain{6.4} & \textbf{23.4}\;\gain{3.2}  \\
    \doubao~\cite{bytedance2025seed16}   & 38.9 & 65.0 & 54.9 & 58.2 & 67.4 & 60.2  & 23.2 \\
    \rowcolor{green!4}
    \methoddb & 32.4 & \textbf{65.9} & \textbf{65.6} & \textbf{63.3} & \textbf{78.0} & \textbf{65.0}\;\gain{4.8} & \textbf{27.5}\;\gain{4.3} \\
    \bottomrule
\end{tabular}
\label{tab:beacon3d_eval}
\vspace{-5pt}
\end{table}

%% file: tables/coherence.tex
\begin{figure}[t]
    \centering

    \begin{minipage}[t]{0.52\linewidth}
        \vspace{0pt}
        \centering
        \captionof{table}{\textbf{Grounding-QA Chain analysis on Beacon3D.} Results are reported for different Grounding-QA Chain evaluation metrics and the overall object-level QA Score.}
        \vspace{10pt}
        \label{tab:coherence}
        \resizebox{\linewidth}{!}{%
        \begin{tabular}{l | c c c c c | c }
        \toprule
        \multirow{2}{*}{\textbf{Method}} &\multicolumn{5}{c|}{\textbf{Coherence Metrics}}&\textbf{QA Score} \\
         & \textbf{T1} $\downarrow$ & \textbf{T2} $\downarrow$ & $\mathbf{R_1}$ $\uparrow$ & $\mathbf{R_2}$ $\downarrow$  & \textbf{DF} $\downarrow$  &\textbf{Obj.}$\uparrow$ \\
        \midrule
        3D-VisTA~\cite{zhu20233d}             & 33.4 & 17.4 & 49.3 & 49.9  & 31.7  &8.1\\
        SceneVerse~\cite{jia2024sceneverse}           & 31.7 & 19.5 & 50.7 & 48.9  & 28.3  &6.6\\
        Chat-Scene~\cite{huang2024chat}           & 24.8 & 25.9 & 44.4 & 56.9  & 29.8  &7.8\\
        SceneCOT~\cite{zhu2024unifying}                 & 24.1 & 16.8 & 58.9 & 41.0  & 34.7  &23.2\\
        \midrule
        \methodgpt
      & 18.3 & 24.5 & 47.2 & 39.2  & 19.3  &23.4\\
        \methoddb  & 19.2 & 23.9 & 53.0 & 37.5  & 17.2  &\textbf{27.5}\\
        \bottomrule
        \end{tabular}%
        }
    \end{minipage}
    \hfill
    \begin{minipage}[t]{0.43\linewidth}
        \vspace{5pt}
        \centering
        \includegraphics[width=\linewidth]{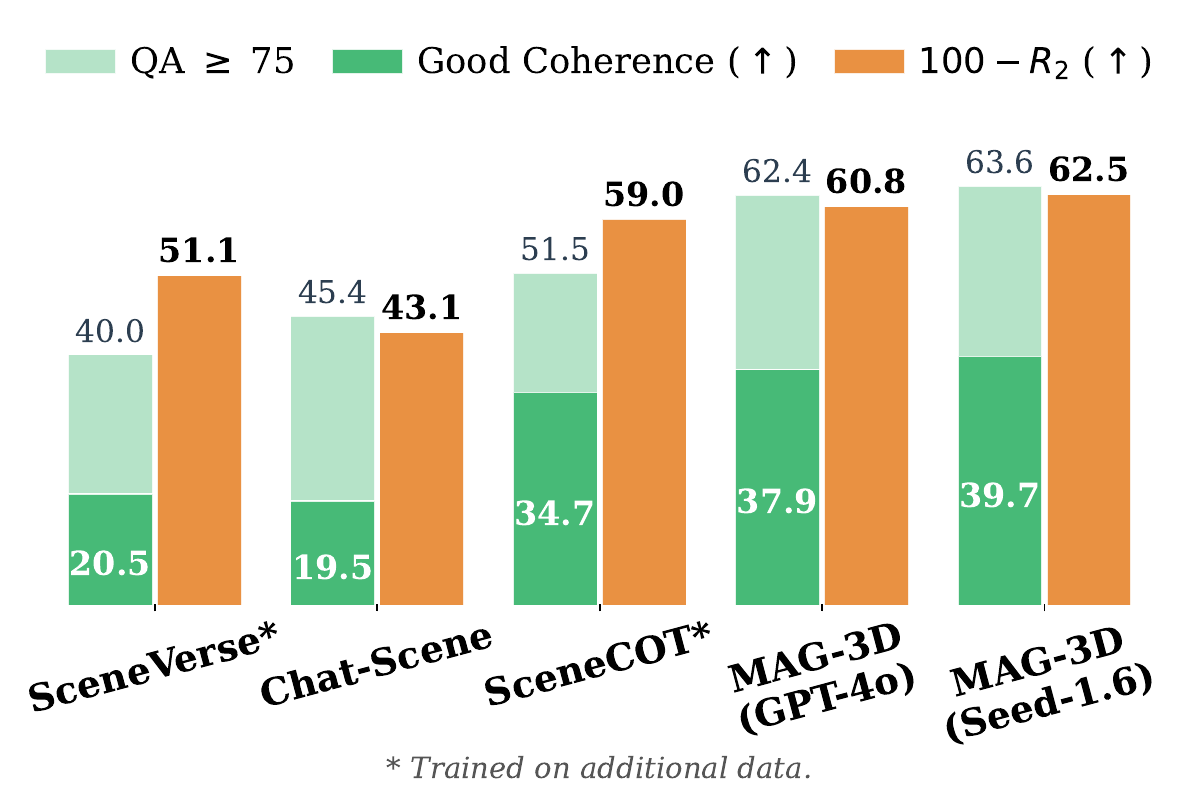}
        \vspace{-20pt}
        \captionof{figure}{\textbf{Grounding-QA coherence on Beacon3D.} Comparison of QA Score $\ge75$, Good Coherence, and $R_2$ across state-of-the-art methods. }
        \label{fig:grounding_qa_coherence}
    \end{minipage}
\vspace{-10pt}
\end{figure}

%% file: tables/msqa.tex
\begin{table}[t]
\centering
\setlength{\tabcolsep}{2.2pt}
\caption{\textbf{Results on MSQA.} Results are reported under the official setting and vision-only setting. T, I, O, and PC denote text, image, object proposals, and point cloud inputs, respectively. The \textcolor{green!60!black}{green} rows represents our methods, it shows that our method consistently improves overall performance across both settings.}
\begin{tabular}{l|c|cccccc|c}
    \toprule
    \multirow{2}{*}{\textbf{Method}} & \multirow{2}{*}{\textbf{Input}}
    & \multicolumn{6}{c|}{\textbf{Question Type}} & \textbf{Overall} \\
    & & \emph{Count.} & \emph{Exist.} & \emph{Attr.} & \emph{Spatial} & \emph{Navi.} & \emph{Others} & \textbf{Score}\\
    \midrule
    \multicolumn{9}{l}{\small\textit{Official Setting}}\\
    LEO~\cite{huang2024embodied}                         & T+PC   & 0.8  & 15.5 & 11.8 & 7.3  & 2.3  & 15.3 & 7.8 \\
    Claude-3.5-Sonnet~\cite{anthropic2024claude35sonnet}           & T+I+O  & 32.6 & 66.3 & 69.9 & 30.1 & 45.5 & 83.6 & 49.7 \\
    GPT-3.5 ~\cite{openai2024chatgptwhisperapis}                     & T+O    & 34.8 & 74.5 & 75.8 & 27.9 & 43.0 & 88.0 & 50.7 \\
    GPT-4o~\cite{hurst2024gpt}                      & T+I+O  & 31.2 & 71.4 & 75.2 & 31.5 & 36.7 & 88.0 & 49.7 \\
    \rowcolor{green!4}
    \methodgpt    & T+I+O  & \textbf{37.0} & \textbf{85.6} & 66.5 & 31.2 & \textbf{53.2} & 79.2 & \textbf{53.0}\;\gain{3.3} \\
    \doubao~\cite{bytedance2025seed16}                  & T+I+O  & 31.3 & 75.1 & 69.4 & 36.8 & 24.0 & 77.1 & 47.9 \\
    \rowcolor{green!4}
    \methoddb    & T+I+O  & \textbf{44.1} & \textbf{88.5} & 62.2 & 33.9 & \textbf{53.5} & 73.7 & \textbf{54.3}\;\gain{6.4} \\
    \midrule
    \multicolumn{9}{l}{\small\textit{Vision-only Setting}}\\
    GPT-4o~\cite{hurst2024gpt}                      & T+I    & 17.9 & 66.5 & 44.1 & 33.4 & 31.1 & 68.4 & 39.4 \\
    \rowcolor{green!4}
    \methodgpt    & T+I    & 17.5 & \textbf{68.7} & \textbf{52.2} & 26.2 & \textbf{38.8} & \textbf{75.4} & \textbf{40.8}\;\gain{1.4} \\
    \doubao~\cite{bytedance2025seed16}                  & T+I    & 15.1 & 62.6 & 26.2 & 22.6 & 21.5 & 51.7 & 29.6 \\
    \rowcolor{green!4}
    \methoddb    & T+I    & \textbf{23.6} & \textbf{70.5} & \textbf{48.8} & \textbf{29.6} & \textbf{42.6} & \textbf{67.0} & \textbf{42.4}\;\gain{12.8} \\
    \bottomrule
\end{tabular}
\label{table:msqa_eval}
\end{table}

%% file: tables/ablation_framework.tex
\begin{table}[t]
\centering
\setlength{\tabcolsep}{3pt}

\begin{minipage}[t]{0.48\textwidth}
    \centering
    \captionof{table}{\textbf{Ablation of proposed components.} \textbf{G.A.}, \textbf{C.A.} denote the grounding agent and coding agent, respectively. While \textbf{T.U.} denotes the non-multi-agent tool-use baseline.}
    \label{tab:ablations_module_core}
    \small
    \begin{tabular}{c|cc|c}
        \toprule
         \multirow{2}{*}{\textbf{T.U.}}& \multicolumn{2}{c|}{\textbf{Multi-Agent}}& \textbf{QA}\\
         & \textbf{G.A.}&\textbf{C.A.}& \textbf{Score} \\
         \midrule
         --& -- & -- & 37.5 \\
         \checkmark& -- & -- & 44.6 \\
         --& $\checkmark$ & -- & 43.6 \\
         --& $\checkmark$ & $\checkmark$ & \textbf{47.6} \\
        \bottomrule
    \end{tabular}
\end{minipage}
\hfill
\begin{minipage}[t]{0.48\textwidth}
    \centering
    \captionof{table}{\textbf{Ablation of agent backbones.} We evaluate alternative backbone choices for the agents; We use the GPT-4o-mini variant for a more manageable cost in ablation experiments.}
    \label{tab:ablations_module_llm}
    \small
    \begin{tabular}{cc|c}
        \toprule
        \textbf{Grounding} & \textbf{Coding} & \textbf{QA}\\
        \textbf{Agent} & \textbf{Agent} & \textbf{Score}\\
        \midrule
        GPT-4o-mini &  Qwen3-Coder & 45.1 \\
        GPT-4o-mini & \doubao & 45.3 \\
        \doubao & Qwen3-Coder & 47.0 \\
        \doubao & \doubao & \textbf{47.6} \\
        \bottomrule
    \end{tabular}
\end{minipage}

\end{table}

%% file: tables/ablation_methods.tex
\begin{table}[t]
\centering
\setlength{\tabcolsep}{2pt}

\begin{minipage}[t]{0.48\textwidth}
    \centering
    \captionof{table}{\textbf{Ablation of grounding methods.} We compare no grounding, Mask3D-based grounding, and our open-vocabulary Grounding Agent (G.A.). \emph{Count.} and \emph{Exist.} are grounding-sensitive categories, and improvements on them correlate with higher QA Score.}
    \label{tab:grounding_comparison}
    \small
    \begin{tabular}{l|cc|c}
        \toprule
        \textbf{Method}& \emph{Count.}& \emph{Exist.}&\textbf{QA Score} \\
        \midrule
        None & 14.1 & 62.5 & 37.5 \\
        Mask3D\cite{Schult23mask3d} & 20.9 & 71.9 & 38.5 \\
        G.A. &\textbf{27.8} & \textbf{78.1}&\textbf{47.6} \\
        \bottomrule
    \end{tabular}
\end{minipage}
\hfill
\begin{minipage}[t]{0.48\textwidth}
    \centering
    \captionof{table}{\textbf{Ablation of visual memory strategies.} “None (Uniform)” uses no explicit visual memory caching and uniformly samples input images, “SAM3” selects frames based on 2D cues, and our 3D-based method retrieves views aligned with the queried 3D region.}
    \label{tab:zoomer_comparison}
    \small
    \begin{tabular}{l|c}
        \toprule
        \textbf{Visual Memory} & \textbf{QA Score} \\
        \midrule
        None (Uniform) & 44.3 \\
        2D-based (SAM3~\cite{carion2025sam3segmentconcepts}) & 46.0 \\
        3D-based (Ours) & \textbf{47.6} \\
        \bottomrule
    \end{tabular}
\end{minipage}
\vspace{-5pt}
\end{table}

%% file: sec/5conclusion.tex
\section{Conclusion}

Grounded 3D reasoning depends on accurate localization of task-relevant entities as well as spatial and geometric inference that remains consistent with the scene’s structure. We introduced \method, a training-free multi-agent framework that tackles this challenge by flexibly coordinating open-vocabulary grounding, visual memory retrieving, and programmatic geometric verification without task-specific training or hand-crafted pipelines. On both Beacon3D and MSQA, \method achieves state-of-the-art performance compared to previous method. Beyond overall accuracy, our Grounding--QA coherence analysis shows that \method demonstrate grounded 3D ability by aligning correct answers with correct grounding, reducing cases where a seemingly correct response is produced despite incorrect instance grounding. Moreover, the design of \method makes it straightforward to incorporate stronger backbones or improved perception modules as they become available, without changing the overall framework. Overall, these results suggest that agentic coordination with explicit grounding and verification offers a practical path toward flexible and dependable 3D reasoning in open-world settings. Looking forward, \method may serve as an automatic annotator to help bootstrap grounded 3D reasoning data at scale, for example by generating object references, evidence-linked views, spatial relations, and verification traces from large collection of readily available data from web, which can then be filtered or validated with human oversight. We hope our findings encourage further work on grounded reasoning for 3D spatial understanding.

%% file: sec/appendix.tex
\newpage
\appendix
\renewcommand\thefigure{A\arabic{figure}}
\setcounter{figure}{0}
\renewcommand\thetable{A\arabic{table}}
\setcounter{table}{0}
\renewcommand\theequation{A\arabic{equation}}
\setcounter{equation}{0}
\pagenumbering{Alpha}
\renewcommand*{\thepage}{A\arabic{page}}
\setcounter{footnote}{0}

\refstepcounter{part}
\addcontentsline{toc}{part}{Appendix}

{\Large\noindent\textbf{Appendix}\par\medskip}

\renewcommand{\thesection}{\Alph{section}}

\etocsetnexttocdepth{subsection}
\etocsettocstyle
  {\noindent\textbf{Contents of this appendix}\par\medskip}
  {\medskip}
\localtableofcontents

\section{Grounding Agent Implementation Details}

\subsection{Open-Vocabulary 3D Grounding Pipeline}
Given a set of target labels, object descriptions, or free-form queries, together with RGB frames and optionally camera poses (e.g., for the MSQA dataset, where pose information is required to align the scene with the pose specified in the input question for situated reasoning), our pipeline constructs a structured scene memory for open-vocabulary 3D grounding. It progressively transforms language-guided 2D observations into geometrically consistent 3D object instances through dense geometry estimation, mask refinement, 2D-to-3D lifting, cross-frame association, and instance consolidation. The final output is a set of detected objects, each represented by a semantic label and an oriented 3D bounding box.

\subsubsection{Dense Geometry Estimation (VGGT).}
For each RGB frame, we apply VGGT to estimate dense scene geometry, yielding both a depth map and a dense point map. Given camera intrinsics $K$ and pixel coordinates $(u,v)$, the corresponding 3D point is obtained by back-projecting the predicted depth:
\begin{equation}
\mathbf{p}(u,v) = D(u,v)\,K^{-1}[u, v, 1]^\top,
\end{equation}
where $D(u,v)$ denotes the predicted depth value at pixel $(u,v)$. We discard invalid points with zero depth or non-finite values.

\subsubsection{Language-Guided 2D Segmentation with Reflection Handling.}
Mirror reflections often introduce false object evidence in indoor scenes, which can adversely affect downstream 3D grounding and reconstruction by producing duplicated instances and geometrically inconsistent associations. This issue is particularly pronounced for language-guided segmentation, where reflected objects may still match the query semantics despite not corresponding to physical scene entities.

We therefore perform category- or description-conditioned instance segmentation using SAM3 and explicitly account for mirrors. For each query prompt, SAM3 returns a binary mask and an associated confidence score. When explicit labels are provided, we additionally query \emph{mirror} masks to detect reflective regions. If mirror masks are present, then for each non-mirror mask $M_i$, we measure its overlap with each mirror mask $M_m$. When
\begin{equation}
\frac{|M_i \cap M_m|}{|M_i|} > 0.5,
\end{equation}
we regard $M_i$ as being primarily explained by reflection and merge it into the corresponding mirror mask $M_m$ during filtering.

\subsubsection{Mask Filtering and Overlap Re-prompting.}
The raw mask proposals frequently exhibit duplicate detections and cross-category overlaps. To address these issues, we adopt a two-stage post-processing pipeline. During initial filtering, we discard masks with low confidence scores, masks covering less than \(0.1\%\) of the image, and masks with excessive boundary coverage, defined as cases where more than \(50\%\) of the mask area lies within a \(5\%\) image margin. We additionally remove duplicate masks within each category using greedy suppression: masks are considered in descending order of area, and any mask with IoU greater than \(0.8\) with an already retained mask is removed, keeping the larger one. Mirror-aware processing is applied subsequently, as described above.

For inter-class overlapping masks, we introduce a \emph{visual re-prompting} step to resolve ambiguous regions. Specifically, we revisit the overlapping area, gather the candidate semantic labels involved, and re-query SAM3 within the local region using each label as a text prompt. The overlap is then assigned to the most confident prediction, while inconsistent mask regions are suppressed accordingly. After this refinement, we further discard masks that become too small or largely occluded.

\subsubsection{2D-to-3D Lifting and Instance Association.}
Given each filtered 2D mask, we lift its pixels into a 3D point set using the VGGT point map and retrieve the corresponding RGB values, normalized to $[0,1]$. As the lifted points may contain noise, outliers, and disconnected fragments, we apply a sequence of geometric refinements, including subsampling, statistical outlier removal, and a fast point cloud clustering. We retain only the largest resulting cluster as the 3D object proposal and discard masks whose reconstructed 3D support is insufficient.

To build consistent object instances across views, we then associate the per-frame 3D clusters with a global instance set. Each instance maintains its accumulated 3D points, a voxelized occupancy representation for efficient overlap computation, an axis-aligned bounding box for fast candidate rejection, and the set of frames in which it is observed. For each new cluster, we first eliminate candidates with non-overlapping bounding boxes, and then compute a voxel-based bidirectional containment score against the remaining instances:
\begin{equation}
s_\text{o}(A,B) = \max\left(\frac{|V_A \cap V_B|}{|V_A|},\ \frac{|V_A \cap V_B|}{|V_B|}\right),
\end{equation}
where $V_A$ and $V_B$ denote the corresponding voxel sets. The cluster is merged into the best-matching instance if the overlap score exceeds a threshold; otherwise, a new instance is initialized. In this way, noisy per-frame segmentations are consolidated into coherent 3D object instances over time.

\subsubsection{3D Instance Refinement and Consolidation.}
After all frames have been processed, we apply a final refinement stage to improve the quality and consistency of the reconstructed instances. We first discard instances with insufficient geometric support and relabel the remaining instances with contiguous global IDs. For each valid instance, we estimate an oriented bounding box by searching over candidate yaw angles in the horizontal plane and selecting the orientation that yields the most compact trimmed bounding area, thereby improving robustness to outliers.

Finally, we perform a category-wise BEV merging step to consolidate fragmented detections. Instances with sufficiently overlapping bird’s-eye-view footprints are merged into a single object hypothesis, with their points and observation histories aggregated accordingly. This post-processing step helps suppress over-segmentation and produces more coherent final 3D object instances.

\subsection{3D Visual Memory Retrieval Mechanism}
This section provides additional details of the retrieval mechanism in our 3D Visual Memory. In particular, we describe how the queried 3D region is instantiated under different query types, while keeping the same memory representation and retrieval criterion. Depending on the reasoning scenario, the memory can be queried either with a grounded 3D instance or with a direction-conditioned local region for situation reasoning.

Our 3D Visual Memory is implemented as a reusable retrieval structure built on top of the per-frame VGGT reconstruction, and supports both object-centric and region-centric view selection. In addition to the RGB image, each memory entry stores the frame-wise dense point map $\mathbf{P}^{(i)}\in\mathbb{R}^{H_p\times W_p\times 3}$ in the global scene coordinate system, enabling later retrieval to directly identify the views that best cover the queried 3D object or region.

\subsubsection{Instance-Conditioned Retrieval.}
For instance-centric retrieval, the queried region is instantiated directly from the grounded 3D box. Concretely, if the memory retrieval query provides instance bounding box,
\[
\mathbf{b}=(x_{\min},y_{\min},z_{\min},x_{\max},y_{\max},z_{\max}),
\]
we define the corresponding axis-aligned cuboid
\[
\mathcal{R}_{\mathrm{ins}}(\mathbf{b})
=
[x_{\min},x_{\max}] \times [y_{\min},y_{\max}] \times [z_{\min},z_{\max}]
\]
as the retrieval region. The cached top-$K$ views therefore correspond to the frames whose reconstructed points provide the largest support within this object region.

\subsubsection{Direction-Conditioned Retrieval.}
For situation-reasoning queries, we additionally define the visual-memory retrieval region using an anchor point and a horizontal direction vector, while retaining the same 3D visual-memory scoring and caching mechanism described above. Given an anchor point $\mathbf{x}_{\mathrm{ref}}$ and a queried direction $\mathbf{r}$, we project the queried direction onto the ground plane and define a local BEV-aligned coordinate frame centered at $\mathbf{x}_{\mathrm{ref}}$. Specifically, we use the normalized front axis
\[
\mathbf{e}_{\mathrm{front}}=\frac{(r_x,r_y,0)}{\|\mathbf{r}\|},
\]
the lateral in-plane axis
\[
\mathbf{e}_{\mathrm{lat}}=(-e_{\mathrm{front},y},\,e_{\mathrm{front},x},\,0),
\]
which is perpendicular to $\mathbf{e}_{\mathrm{front}}$ in the ground plane, and the world vertical axis
\[
\mathbf{e}_{\mathrm{up}}=(0,0,1).
\]
We then place a front-facing local cuboid in front of $\mathbf{x}_{\mathrm{ref}}$, aligned with this ground-plane coordinate frame, with a $3\mathrm{m}\times 3\mathrm{m}$ footprint in BEV:
\[
\mathcal{R}_{\mathrm{front}}(\mathbf{x}_{\mathrm{ref}},\mathbf{r})
=
\left\{
\mathbf{x}\in\mathbb{R}^3~
\middle|\,
\begin{aligned}
&0 \le \langle \mathbf{x}-\mathbf{x}_{\mathrm{ref}},\mathbf{e}_{\mathrm{front}}\rangle \le 3,\quad
\left|\langle \mathbf{x}-\mathbf{x}_{\mathrm{ref}},\mathbf{e}_{\mathrm{lat}}\rangle\right| \le 1.5,\\
&\langle \mathbf{x}-\mathbf{x}_{\mathrm{ref}},\mathbf{e}_{\mathrm{up}}\rangle \ge 0.1
\end{aligned}
\right\}.
\]
Intuitively, this region captures the local space in front of the queried direction. The visual memory then retrieves frames whose reconstructed 3D points best cover this region. In this way, the point-map memory mechanism remains unchanged, and only the retrieval differs from the instance-centric retrieval used for grounded objects.

\section{Extended Baselines}
\subsection{2D-based Visual Memory Baseline}
As a 2D-based alternative to our 3D visual memory, we also implement a SAM3-based visual memory, whose performance is reported in the ablation study in Tab.~7 of the main paper. Given a query from the Grounding Agent, SAM3 identifies 2D instances that are likely relevant to the query and scores frames based on the presence and spatial extent of these instances. We then cache the top-$K$ ranked frames as the visual memory for downstream reasoning. Compared with the naive no-memory variant, in which each agent must traverse all frames whenever visual evidence is needed, this strategy provides more informative visual context by prioritizing frames that are more likely to contain the queried object. However, because the selection is defined purely in 2D image space, it cannot determine whether the queried object is fully observed across viewpoints, whether the selected frames provide complementary coverage, or whether the retrieved evidence is well aligned with the underlying 3D region. Consequently, although the SAM3-based memory improves over the no-memory baseline, it remains less effective than our 3D-based visual memory for occlusion-aware and region-grounded reasoning.

\subsection{Tool-Use Enhanced LLM Baseline.}
To isolate the effect of explicit multi-agent coordination, we additionally construct a \emph{Tool-Use Enhanced LLM} baseline, whose performance is reported in the component ablation in Tab.~4 of the main paper. Specifically, we remove all sub-agents and attach the same tool set directly to the main Planning Agent, including the grounding module. The model is prompted with the same question as in our full framework, together with a fixed tool-calling template that specifies how to invoke tools and how to summarize their outputs. Under this design, a single VLM is responsible for planning, tool selection, evidence gathering, and final answer generation within one unified context, without explicit role decomposition or inter-agent delegation. This baseline controls for the benefit of \emph{tool access} itself, so that the comparison with our full system more directly reflects the value of specialized sub-agents and structured coordination. In practice, although the Tool-Use Enhanced LLM already benefits from external grounding and geometric tools, it must handle all decisions within a single reasoning stream, making it less effective than our multi-agent design at organizing intermediate evidence and performing grounded reasoning reliably.

\section{More Qualitative Results}
\begin{figure}[ht]
    \centering 
    \includegraphics[width=1.0\textwidth]{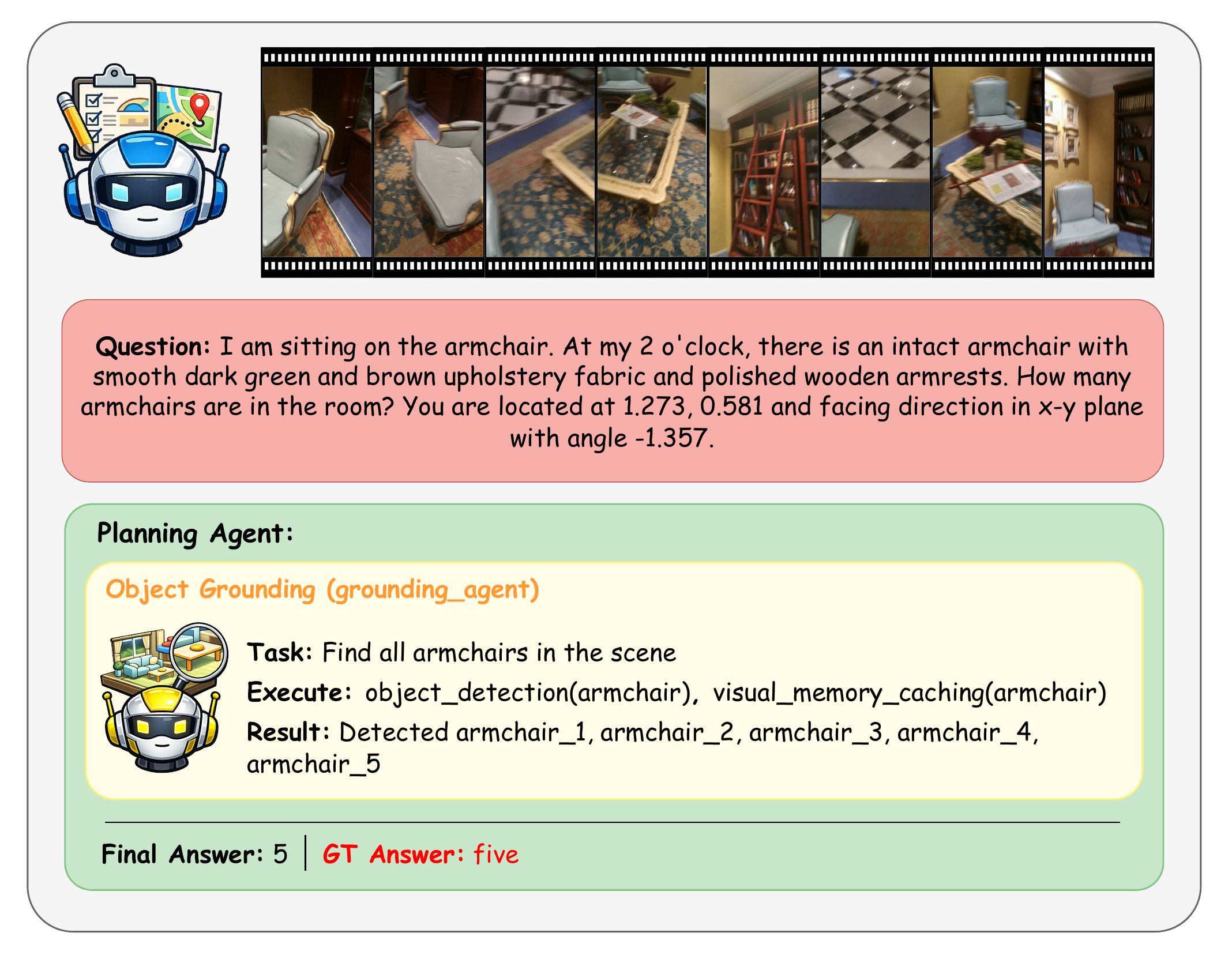}
    \caption{\textbf{Additional qualitative examples on MSQA (Part I).}}
    \label{fig:examples_1}
\end{figure}

\begin{figure}[ht]
    \centering 
    \includegraphics[width=1.0\textwidth]{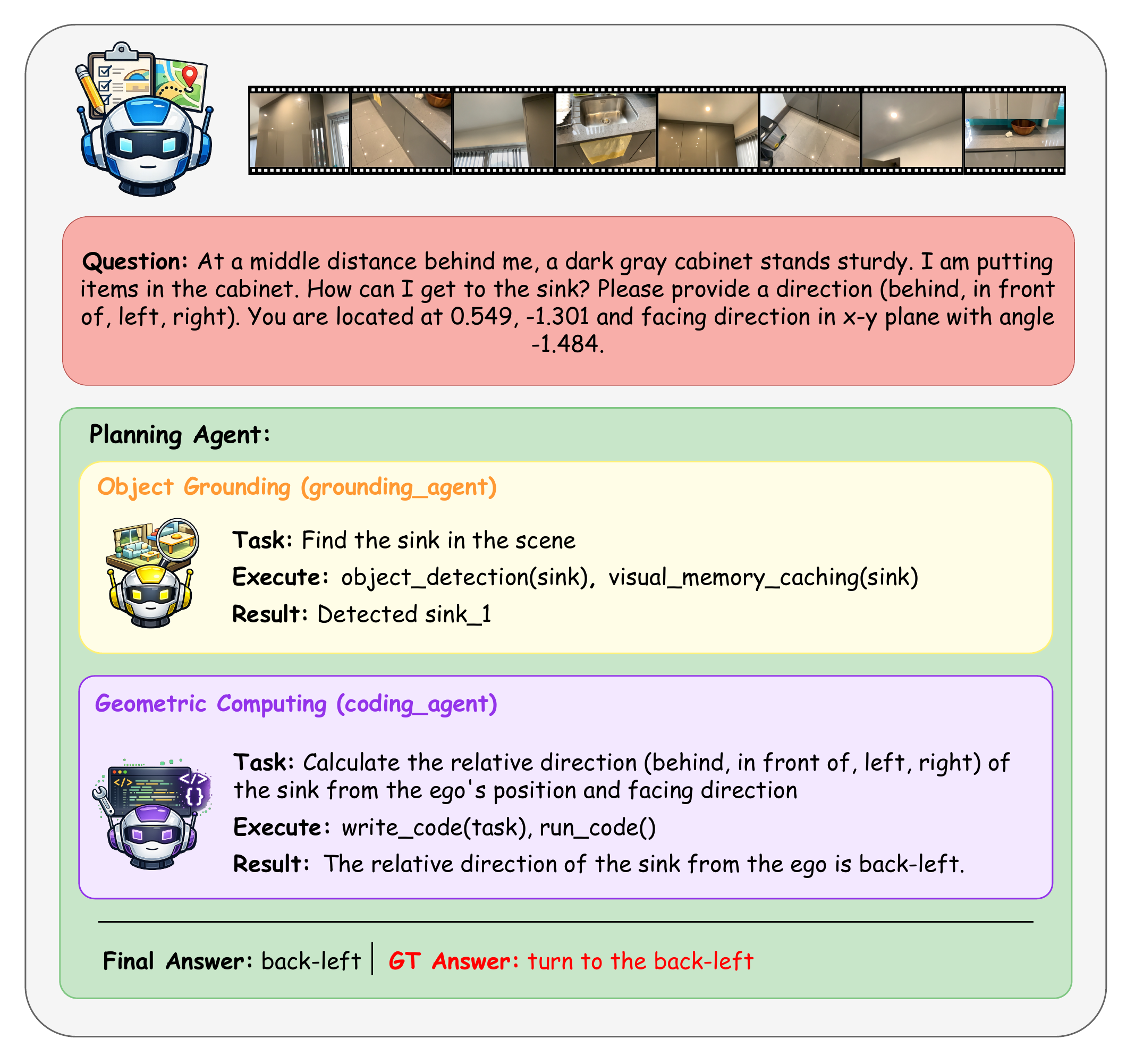}
    \caption{\textbf{Additional qualitative examples on MSQA (Part II).}}
    \label{fig:examples_2}
\end{figure}

\begin{figure}[ht]
    \centering 
    \includegraphics[width=1.0\textwidth]{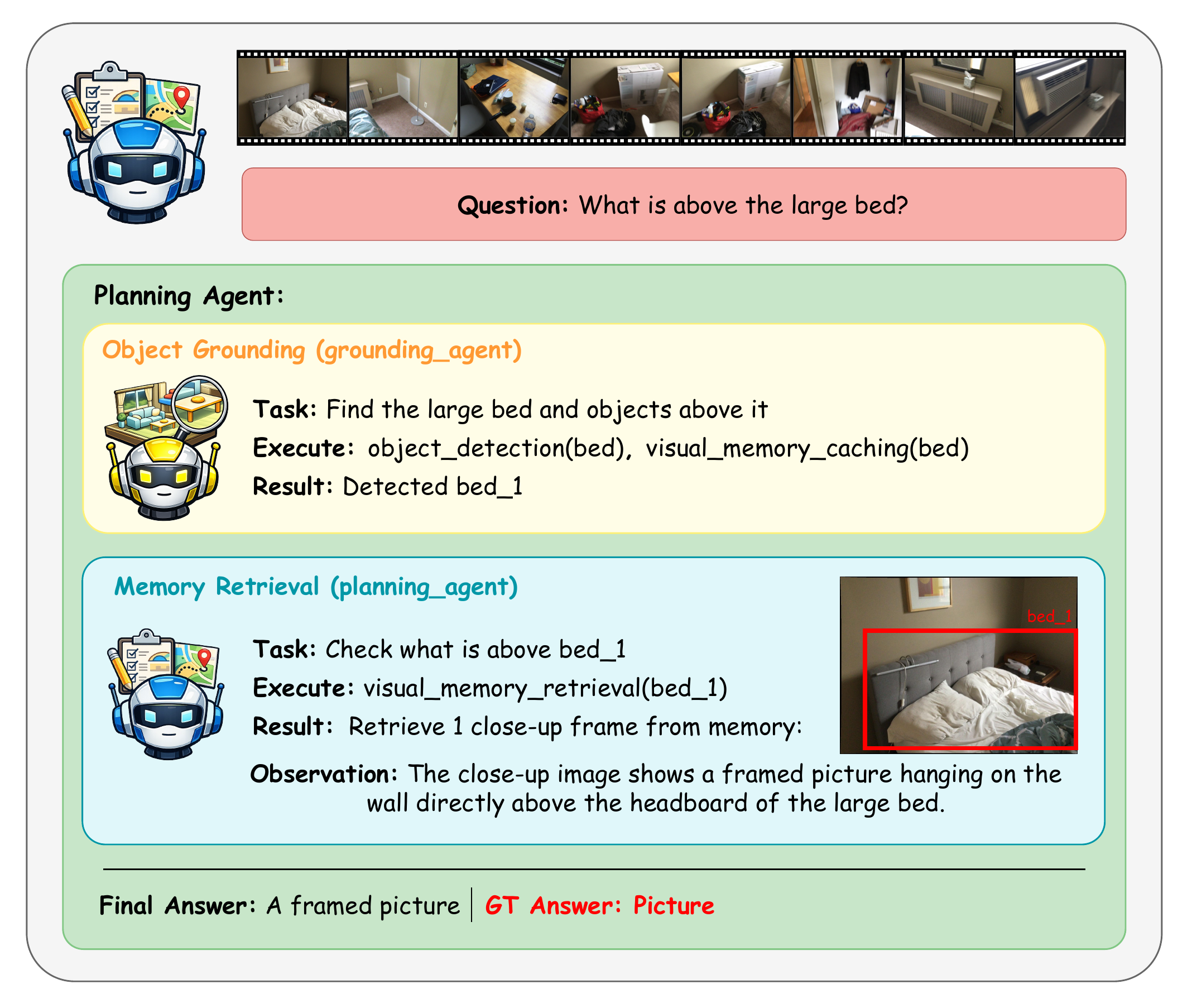}
    \caption{\textbf{Additional qualitative examples on Beacon3D (Part I).}}
    \label{fig:examples_3}
\end{figure}

\begin{figure}[ht]
    \centering 
    \includegraphics[width=1.0\textwidth]{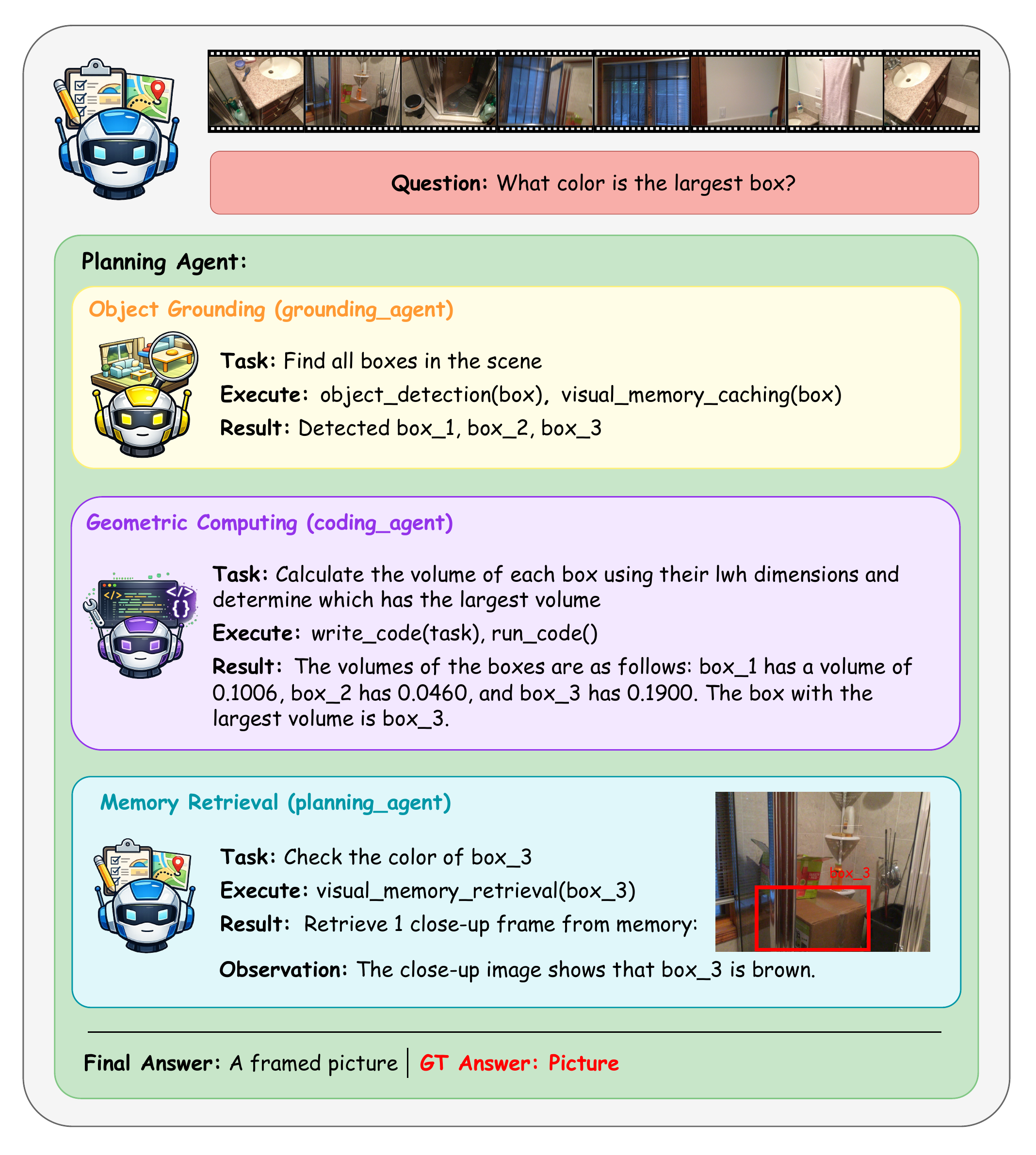}
    \caption{\textbf{Additional qualitative examples on Beacon3D (Part II).}}
    \label{fig:examples_4}
\end{figure}

Figs.~\ref{fig:examples_1} and \ref{fig:examples_2} present additional qualitative examples on Beacon3D, while Figs.~\ref{fig:examples_3} and \ref{fig:examples_4} show additional results on MSQA. These examples further show that \method adapts its reasoning process to each query. Depending on the question, it may focus on object grounding, geometric reasoning or visual inspection. This flexible behavior helps \method handle diverse grounded 3D question-answering scenarios across datasets.

\clearpage
\subsection{MSQA}

\clearpage
\subsection{Beacon3D}
